\documentclass[conference]{IEEEtran}
\IEEEoverridecommandlockouts
\usepackage{cite}
\usepackage{amsmath,amssymb,amsfonts}
\usepackage{algorithmic}
\usepackage{graphicx}
\usepackage{textcomp}
\usepackage{xcolor}
\newcommand{\monodepth}{MonoDepth}
\newcommand{\etal}{et al.}
\usepackage{enumitem}
\setlist[itemize]{nosep}
\usepackage{booktabs}

\def\BibTeX{{\rm B\kern-.05em{\sc i\kern-.025em b}\kern-.08em
		T\kern-.1667em\lower.7ex\hbox{E}\kern-.125emX}}
\begin{document}

\title{How do neural networks see depth in single images?}

\author{\IEEEauthorblockN{Tom van Dijk}
	\IEEEauthorblockA{\textit{Micro Air Vehicle laboratory} \\
		Faculty of Aerospace Engineering\\
		Technische Universiteit Delft\\
		Delft, The Netherlands\\
		{\tt\small J.C.vanDijk-1@tudelft.nl}}
	\and
	\IEEEauthorblockN{Guido de Croon}
	\IEEEauthorblockA{\textit{Micro Air Vehicle laboratory} \\
		Faculty of Aerospace Engineering\\
		Technische Universiteit Delft\\
		Delft, The Netherlands\\
		{\tt\small G.C.H.E.deCroon@tudelft.nl}}
}

\maketitle

\begin{abstract}
Deep neural networks have lead to a breakthrough in depth estimation from single images. Recent work often focuses on the accuracy of the depth map, where an evaluation on a publicly available test set such as the KITTI vision benchmark is often the main result of the article. While such an evaluation shows how well neural networks can estimate depth, it does not show \emph{how} they do this. To the best of our knowledge, no work currently exists that analyzes what these networks have learned.

In this work we take the \monodepth\ network by Godard et al. and investigate what visual cues it exploits for depth estimation.
We find that the network ignores the apparent size of known obstacles in favor of their vertical position in the image.
Using the vertical position requires the camera pose to be known; however we find that \monodepth\ only partially corrects for changes in camera pitch and roll and that these influence the estimated depth towards obstacles.
We further show that \monodepth's use of the vertical image position allows it to estimate the distance towards arbitrary obstacles, even those not appearing in the training set, but that it requires a strong edge at the ground contact point of the object to do so. In future work we will investigate whether these observations also apply to other neural networks for monocular depth estimation.
\end{abstract}

\section{Introduction}



When depth has to be estimated from camera images, stereo vision is a common choice. When only a single camera can be used, optical flow can provide a measure of depth; or if images can be combined over longer time spans then SLAM or Structure-from-Motion can be used to estimate the geometry of the scene. These methods treat depth estimation as a purely geometrical problem, generally completely ignoring the \emph{content} of the images.

When only a \emph{single} image is available, it is not possible to use epipolar geometry. Instead, algorithms have to rely on \emph{pictorial cues}: cues that indicate depth within a single image, such as the apparent size of known objects. Pictorial cues require knowledge of the environment, which makes them difficult to program. As a result, pictorial cues have seen relatively little use in computer vision until recently.

With the arrival of stronger hardware and better machine-learning techniques -- most notably Convolutional Neural Networks (CNN) -- it is now possible to \emph{learn} pictorial cues from data rather than program them by hand.
One of the earliest examples of monocular depth estimation using machine learning was published in 2006 by Saxena \etal \cite{Saxena2006}. In 2014, Eigen \etal \cite{Eigen2014} were the first to use a CNN for monocular depth estimation. Where \cite{Eigen2014} still required a true depth map for training, in 2016 Garg \etal proposed a new scheme that allows the network to learn directly from stereo pairs instead \cite{Garg2016}; this work was further improved upon by Godard \etal in \cite{Godard2017}. Since then, the interest in monocular depth estimation has rapidly grown, with new articles appearing on a weekly basis during the past year.

Recent work has focused primarily on improving the accuracy of monocular depth estimation, where evaluations on publicly available datasets such as KITTI \cite{Menze2015} and NYUv2 \cite{Silberman2012} show that computers \emph{can} indeed estimate depth from single images.
However, to the best of our knowledge, no work exists that investigates \emph{how} they do this.

Why is it important to know what these neural networks have learned? First, it is difficult to guarantee correct behavior without knowing what the network does. Evaluation on a test set shows that it works correctly in those cases, but it does not guarantee correct behavior in other unexpected scenarios. Second, knowing what the network has learned provides insight into training. Additional guidelines for the training set and data augmentation might be derived from the learned behavior.
Third, it provides insight into transfer to other setups. With an understanding of the network, it is for instance easier to predict what the impact of a change in camera height will be and whether this will work out-of-the-box, require more data augmentation or even a new training set.

In this work we take the \monodepth\ network by Godard \etal \cite{Godard2017,Godard2017_github} and investigate its \emph{high-level behavior}. Section \ref{sec:relatedwork} gives an overview of related work on monocular depth estimation in humans and on the analysis of neural networks. In Section \ref{sec:pos_vs_scale} we show that \monodepth\ primarily relies on the vertical image position of obstacles but not on their apparent size. Using the vertical position requires knowledge of the camera pose; in Section \ref{sec:flatground} we investigate whether the camera pose is assumed constant based on the training set or is observed from the test images. In Section \ref{sec:appearance} we investigate how \monodepth\ separates obstacles from background and what factors allow it to detect objects not appearing in the training set. We discuss the impact of our results in Section \ref{sec:discussion}.

\section{Related work}
\label{sec:relatedwork}

Existing work on monocular depth estimation has extensively shown that neural networks can estimate depth from single images, but an analysis of how this estimation works is still missing. That does not mean that no analysis methods for neural networks exist. Feature visualization and attribution can be used to analyze the behavior of CNNs. One of the earlier examples of feature visualization in deep networks can be found in \cite{Erhan2009}. The methods have been improved upon in e.g. \cite{Szegedy2013,Zeiler2013} and an extensive treatment of visualization techniques can be found in \cite{Olah2017a}. In essence, the features used by a neural network can be visualized by optimizing the input images with respect to a loss function based on the excitation of a single neuron, a feature map or an entire layer of the network.
While these methods provide insight into the workings of a CNN (and their use is absolutely recommended for future work), they do not directly answer the question how neural networks estimate depth. A collection of features that the neural network is sensitive to is not an explanation of its behavior; further interpretation and experiments are still required.

In this work, we take a different approach that is perhaps more closely related to the study of (monocular) depth perception in humans. We treat the neural network as a black box, only measuring the responses (in this case depth maps) to certain inputs. Rather than optimizing the inputs with regards to a loss function, we modify or disturb the images, for instance by adding conflicting visual cues, and look for a correlation in the resulting depth maps. 

Literature on human depth perception provides insight into the pictorial cues that could be used to estimate distance.
The following cues from \cite{Gibson1950} and more recent reviews \cite{Cutting1995,Brenner2018} can typically be found in single images:
\begin{itemize}
    \item Position in the image. Objects that are further away tend to be closer to the horizon. When resting on the ground, the objects also appear higher in the image.
    \item Occlusion. Objects that are closer occlude those that lie behind them. Occlusion provides information on depth order, but not distance.
    \item Texture density. Textured surfaces that are further away appear more fine-grained in the image.
    \item Linear perspective. Straight, parallel lines in the physical world appear to converge in the image.
    \item Apparent size of objects. Objects that are further away appear smaller.
    \item Shading and illumination. Surfaces appear brighter when their normal points towards a light source. Light is often assumed to come from above. Shading typically provides information on depth changes within a surface, rather than relative to other parts of the image.
    \item Focus blur. Objects that lie in front or behind the focal plane appear blurred.
    \item Aerial perspective. Very far away objects (kilometers) have less contrast and take on a blueish tint.
\end{itemize}

Of these cues, we expect that only the position in the image and apparent size of objects are applicable to the KITTI dataset; other cues are unlikely to appear because of low image resolution (texture density, focus blur), limited depth range (aerial perspective) or they are less relevant for distance estimation towards obstacles (occlusion, linear perspective and shading and illumination).


Both cues have been experimentally observed in humans, also under conflicting conditions. Especially the vertical position in the visual field has some important nuances. For instance, Epstein shows that perceived distances do not solely depend on the vertical position in the visual field, but also on the background \cite{Epstein1966a}. With a ceiling-like background, objects lower in the visual field are perceived further away instead of the other way round.
Another important contextual feature is the horizon line, which gains further importance when little ground (or ceiling) texture is present \cite{Gardner2010}. Using prismatic glasses that manipulated the human subjects' vision, Ooi \etal showed that humans in real-world experiments use the angular declination relative to the `eye level' \cite{Ooi2001} rather than the visual horizon. The eye level is the \emph{expected} height of the horizon in the visual field. 

The apparent size of objects also influences their estimated distance. Sousa \etal performed an experiment where subjects needed to judge distances to differently-sized cubes \cite{Sousa2012}. The authors observed that the apparent size of the cubes influenced the estimated distance even though the true size of the cubes was not known and the height in the visual field and other cues were present.

No work was found that investigates whether the observations described above also apply to neural networks for depth estimation.

\begin{figure}
    \centering
    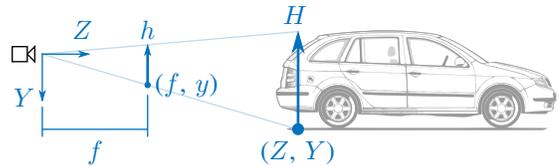
    \caption{True object size $H$ and position $Y$, $Z$ in the camera frame, vertical image position $y$ and apparent size $h$. Image coordinates (e.g. $y$) are measured from the center of the image.}
    \label{fig:depth_cues}
\end{figure}

\begin{figure*} 
    \centering
    \includegraphics{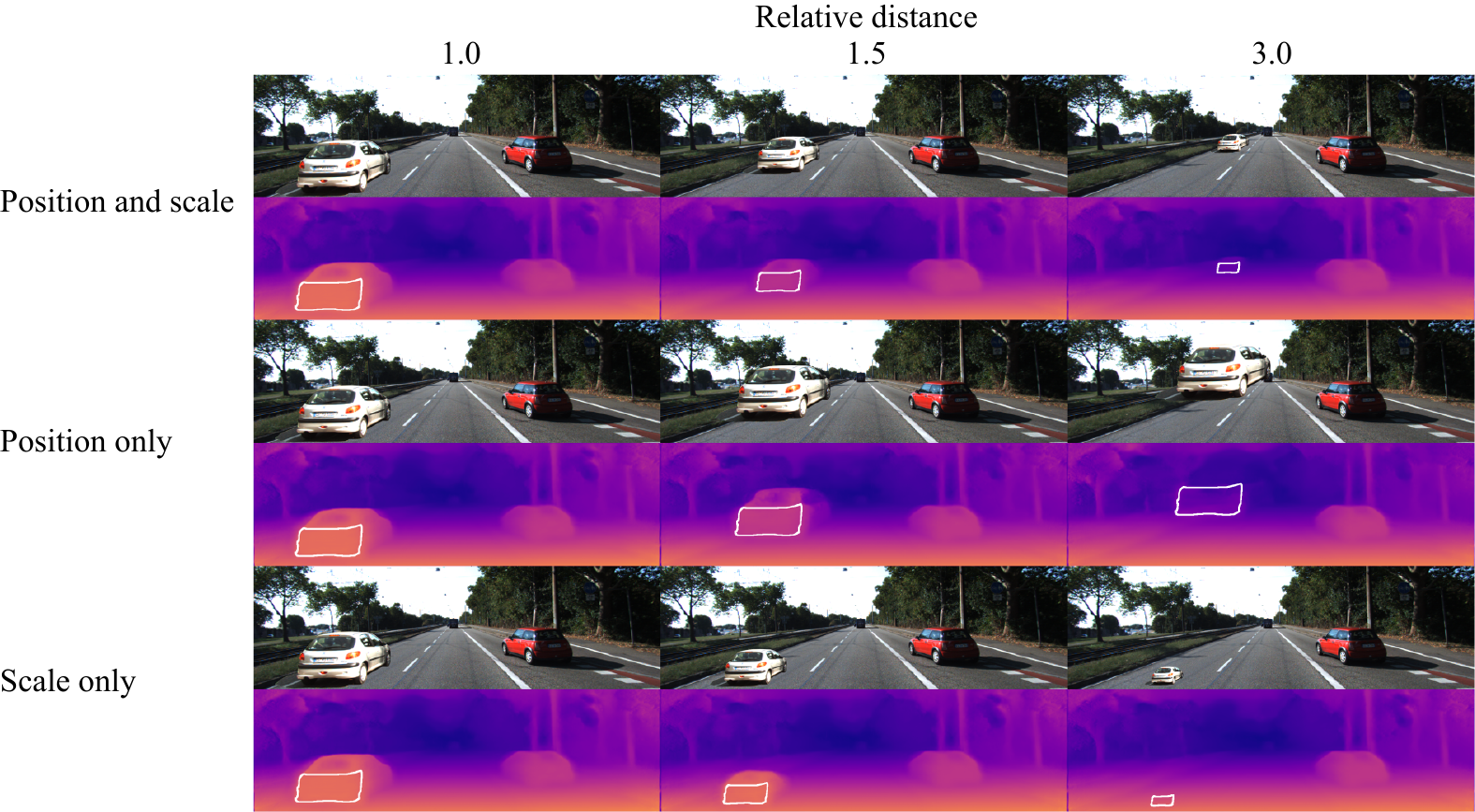}
    \caption{Example test images and resulting disparity maps. The white car on the left is inserted into the image at a relative distance of 1.0 (left column), 1.5 (middle column) and 3.0 (right column), where a distance of 1.0 corresponds to the same scale and position at which the car was cropped from its original image. In the top row, both the position and scale of the car vary with distance, in the middle row only the position changes and the scale is kept constant, and in the bottom row the scale is varied while the position remains constant. The measurement region from which the estimated distance is obtained is indicated by a white outline in the disparity maps.}
    \label{fig:pos_vs_scale_examples}
\end{figure*}

\begin{figure}
    \centering
    \includegraphics{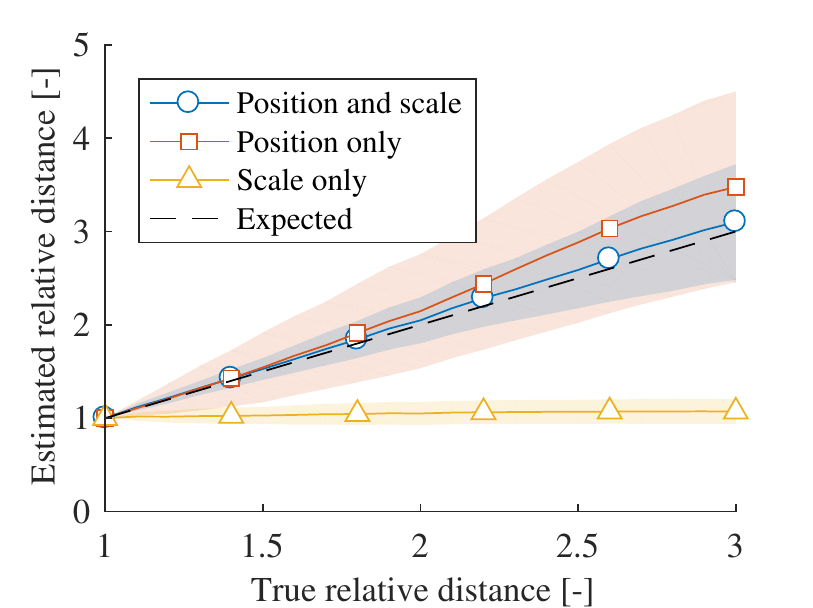}
    \caption{Influence of vertical image position and apparent size on the depth estimate. 
    Shaded regions indicate $\pm 1\,\text{SD}\ (N=1862)$. When both depth cues are present, the network can successfully estimate the distance towards the object. When only the vertical position is available, the distance is slightly overestimated and the standard deviation of the measurement increases. When only scale information is available, the network is no longer able to estimate distance.}
    \label{fig:pos_vs_scale}
\end{figure}

\section{Position vs. apparent size}
\label{sec:pos_vs_scale}
As stated in Section \ref{sec:relatedwork}, the vertical image position and apparent size of objects are the most likely cues to be used by the network.
Figure \ref{fig:depth_cues} shows how these cues can be used to estimate the distance towards obstacles. The camera's focal length is assumed known and constant, and is learned implicitly by the neural network. Given the obstacle's real-world size $H$ and apparent size $h$ in the image, the distance can be estimated using:
\begin{equation}
    Z = \frac{f}{h} H
\end{equation}
This requires the obstacle's true size $H$ to be known. The objects encountered most often in the KITTI dataset come from a limited number of classes (e.g. cars, trucks, pedestrians), where all objects within a class have roughly the same size.
It is therefore possible that the network has learned to recognize these objects and uses their apparent size to estimate their distance.

Alternatively, the network could use the vertical image position $y$ of the object's ground contact point to estimate depth.
Given the height $Y$ of the camera above the ground, the distance can be estimated through:
\begin{equation}
    Z = \frac{f}{y} Y
\end{equation}
This method does not require any knowledge about the true size of the object, but instead assumes the presence of a flat ground and fixed or known camera pose.
These assumptions are also valid in the KITTI dataset.

\subsection{Evaluation method}
To investigate which of these cues are used by the network, three sets of test images are generated: one in which the apparent size of objects is varied but the vertical position of the ground contact point in the image is kept constant. One in which the vertical position is varied but the size remains constant, and finally a control set in which both the apparent size and position are varied with distance -- as would be expected in real-world images.

The test images were generated as follows: the objects (mostly cars) are cropped from the images of KITTI's scene flow dataset. Each object is labeled with its location relative to the camera (e.g. one lane to the left, facing towards the camera) and with its position in the image it was cropped from.
Secondly, each image in the test set was labeled with positions where an obstacle could be inserted (e.g. the lane to the left of the camera is still empty). Combining this information with the object labels ensures that the test images remain plausible.

The true distance to the inserted objects is not known (although it could be extracted from the LIDAR images); instead the network's ability to measure relative distances will be evaluated. The distances are relative to the size and position at which the object was cropped from the image in which it originally appeared, which is assigned a relative distance $Z'/Z=1.0$. The distance is increased in steps of 0.1 up to 3.0 and controls the scaling $s$ and position $x'$, $y'$ of the object as follows:
\begin{equation}
    s = \frac{Z}{Z'},
\end{equation}
and
\begin{equation}
    x' = x \frac{Z}{Z'},\quad y' = h_y + (y - h_y)\frac{Z}{Z'}
\end{equation}
with $x'$, $y'$ the new coordinates of the ground contact point of the (scaled) object and with $h_y$ the height of the horizon in the image which is assumed constant throughout the dataset. The image coordinates $x$, $x'$, $y$, $y'$ and $h_y$ are measured from the center of the image.

The estimated depth towards the car is evaluated by averaging the depth map over a flat region on the front or rear of the car around the front- or taillights (Figure \ref{fig:pos_vs_scale_examples}). A flat region is used rather than the entire object, to prevent the estimated length of the vehicle from influencing the depth estimate; the length is very likely dependent on the apparent size of the object, while the distance might not be.

\subsection{Results}
The results of this experiment are shown in Figure \ref{fig:pos_vs_scale}.
When both the position and scale are varied, the depth estimate behaves as expected: the estimate remains close to the true depth of the object which shows that the network works correctly on these artificial images.
When only the vertical position is varied, \monodepth\ is still able to estimate the distance towards the object, although this distance is slightly overestimated.
Additionally, the standard deviation of the distance estimate has increased significantly compared to the control set.
The most surprising result is found when only the apparent size of the object is changed but the ground contact point is kept constant: \monodepth\ does not observe any change in distance under these circumstances.

These results suggest that \monodepth\ relies primarily on the vertical position of objects rather than their apparent size, although some change in behavior is observed when the size information is removed.
The use of vertical position implies that \monodepth\ assumes the presence of a \emph{flat ground} and has some knowledge of the \emph{camera's pose}.
The next section will investigate these claims more closely.

\section{Camera pose: constant or estimated?}
\label{sec:flatground}

Using the vertical image position to estimate depth towards obstacles requires knowledge of the camera's height and pitch.
There are multiple ways in which the \monodepth\ network can obtain this knowledge: it can find the camera's pitch from the input image (for instance, by finding the horizon or vanishing points), or it can assume that the camera pose is constant.
This assumption should work reasonably well on the KITTI dataset, where the camera is fixed rigidly to the car and the only deviations come from pitch and heave motions from the car's suspension and from slopes in the terrain.
It would, however, also mean that the trained network cannot be directly transfered to another vehicle with a different camera setup.
It is therefore important to investigate whether \monodepth\ assumes a fixed camera pose or can estimate and correct this on-the-fly.

\subsection{Camera pitch}
\begin{figure}
    \centering
    \includegraphics{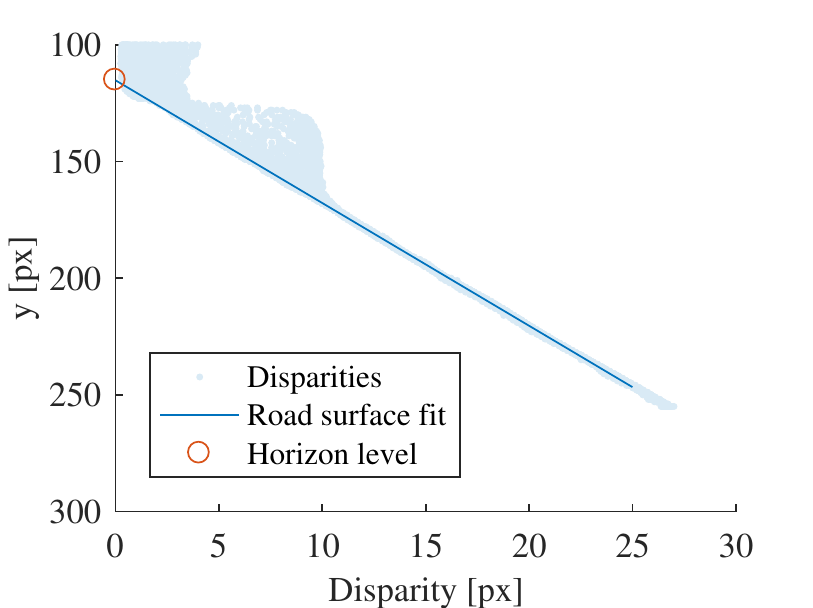}
    \caption{Estimation of the horizon level from true or estimated depth maps. Pixel $y$ and disparity values are collected from the bottom center of the image. The road surface is fitted using RANSAC, the resulting line is then evaluated at a disparity of 0 (i.e. infinite distance) to find the horizon level.}
    \label{fig:pitch_fit}
\end{figure}

If \monodepth\ can measure the camera pitch, then changes in the camera pitch should also be reflected in the estimated depth map. If instead it assumes a fixed camera pose above the ground, then the ground surface should remain constant in the depth estimates.
The unmodified KITTI test images already have some variation in the position of the horizon, caused by pitch and heave motions of the car. To test whether \monodepth\ can correct for these changes, we looked for a correlation between the true horizon level in the images (determined from the Velodyne data) and the estimated horizon level in the depth estimate. The horizon levels were measured by cropping a central region of the depth map (the road surface) and using RANSAC to fit a line to the disparity-y pairs (Figure \ref{fig:pitch_fit}). Extrapolating this line to a disparity of zero (i.e. infinite distance) gives the elevation of the horizon in the image. For each image, this procedure was repeated five times to average out fitting errors from the RANSAC procedure.

\begin{figure}
    \centering
    \includegraphics{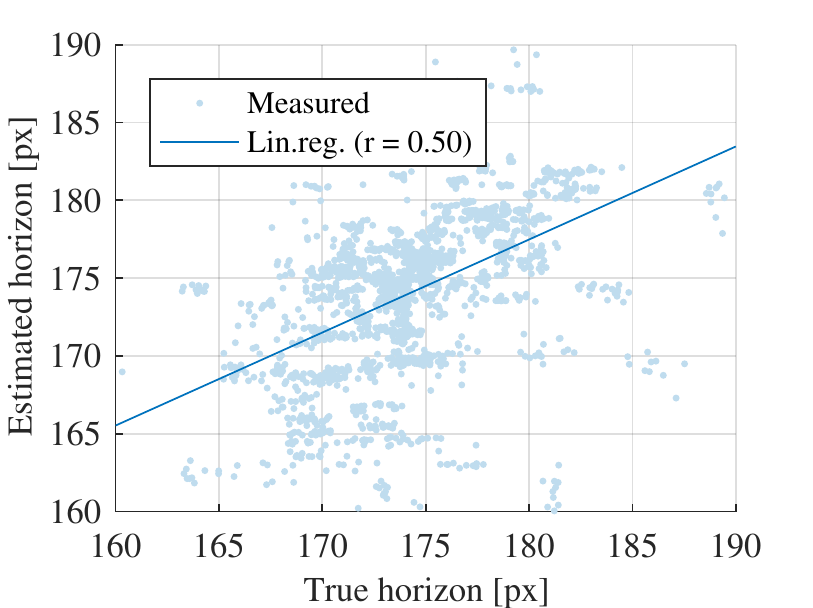}
    \caption{True and estimated horizon levels in unmodified KITTI images. A medium-to-large correlation is found (Pearson's $r=0.50$, $N=1892$) but the slope is only 0.60, indicating that the true shift in the horizon is not fully reflected in the estimated depth map.}
    \label{fig:pitch_unmodified}
\end{figure}

Figure \ref{fig:pitch_unmodified} shows the relation between the true and estimated horizon levels. 
While it was expected that \monodepth\ would either correct for the horizon level or not, a regression coefficient of 0.60 was found which indicates that it does something inbetween these extremes.

\begin{figure}
    \centering
    \includegraphics{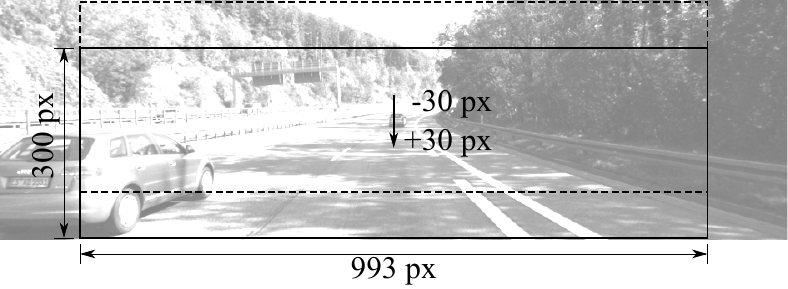}
    \caption{Larger camera pitch angles are emulated by cropping the image at different heights.}
    \label{fig:pitch_crop}
\end{figure}

A second experiment was performed to rule out any issues with the Velodyne data and with the small ($\pm 10\,\mathrm{px}$) range of true horizon levels in the first experiment.
In this second experiment, a smaller region is cropped at different heights in the image (Figure \ref{fig:pitch_crop}). For each image, seven crops are made with offsets between -30 and 30 pixels from the image center, which approximates a change in camera pitch of $\pm$2-3 degrees.
Instead of using the Velodyne data to estimate the true horizon level, the horizon level from the depth estimate of the centrally cropped image is used as a reference value. In other words, this experiment evaluates how well a \emph{shift} in the horizon level is reflected in the depth estimate, rather than its absolute position. 

\begin{figure}
    \centering
    \includegraphics{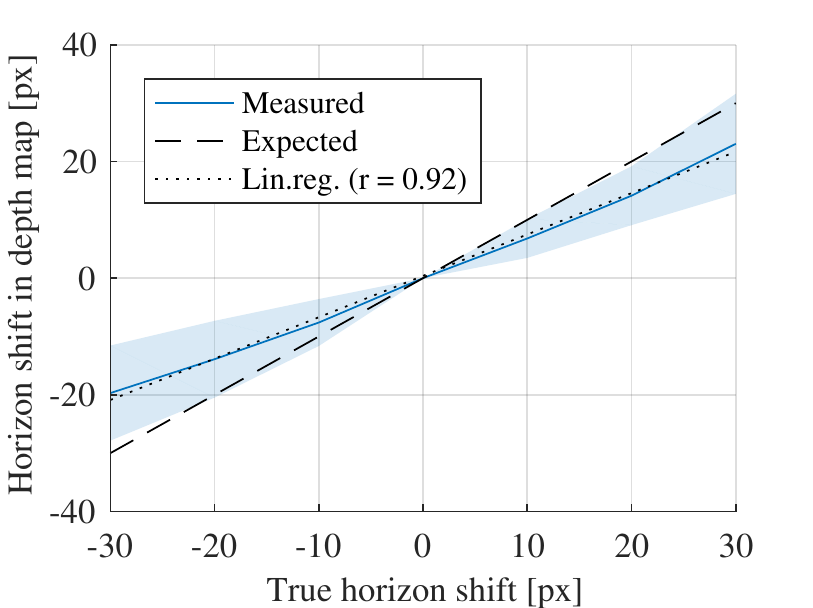}
    \caption{True and estimated shifts in horizon levels after cropping the images at different heights. Shaded regions indicate $\pm 1\,\text{SD}$ ($N=194$, six outliers $>3\,\text{SD}$ removed). A regression coefficient of 0.71 was found (Pearson's $r=.92$, $N=1358$), again indicating that changes in camera pitch do not fully show up in the depth map.}
    \label{fig:pitch}
\end{figure}

The results are shown in Figure \ref{fig:pitch}. A similar result as in the previous experiment is found: \monodepth\ is able to detect changes in camera pitch, but does not fully account for these in the depth estimate. In this experiment a regression coefficient of 0.71 was found.

\begin{figure}
    \centering
    \includegraphics{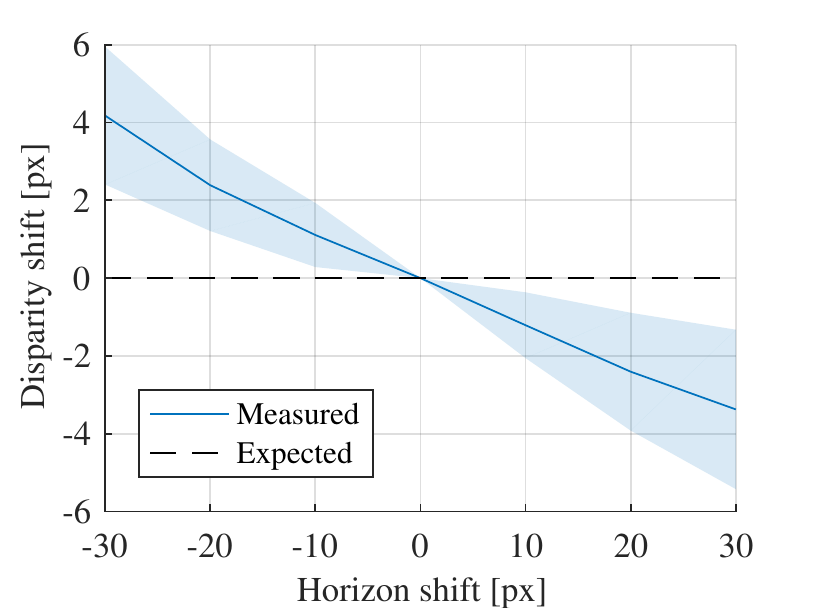}
    \caption{Changes in camera pitch disturb the estimated distance towards obstacles. Shaded regions indicate $\pm 1\,\text{SD}$.}
    \label{fig:pitch_vs_disp}
\end{figure}

Since \monodepth\ uses the vertical position of obstacles to estimate depth, we expect this failure to fully correct for the camera pitch to affect the estimated distances.
To test this hypothesis, we use the same pitch crop dataset and evaluate whether a change in camera pitch causes a change in obstacle disparities. The obstacles were selected by hand from the 200 training images in the KITTI scene flow dataset, where the obj\_map images were used to select the pixels belonging to these obstacles.
The results are shown in Figure \ref{fig:pitch_vs_disp}. 
The distance to obstacles is indeed affected by camera pitch.

\subsection{Camera roll}

\begin{figure}
    \centering
    \includegraphics{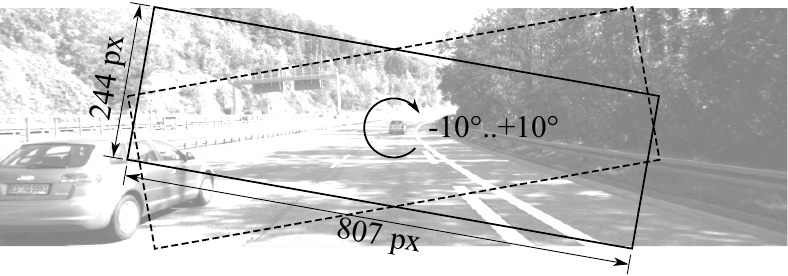}
    \caption{Camera roll angles are emulated by cropping smaller, tilted regions from the original KITTI images.}
    \label{fig:roll_crop}
\end{figure}

\begin{figure}
    \centering
    \includegraphics{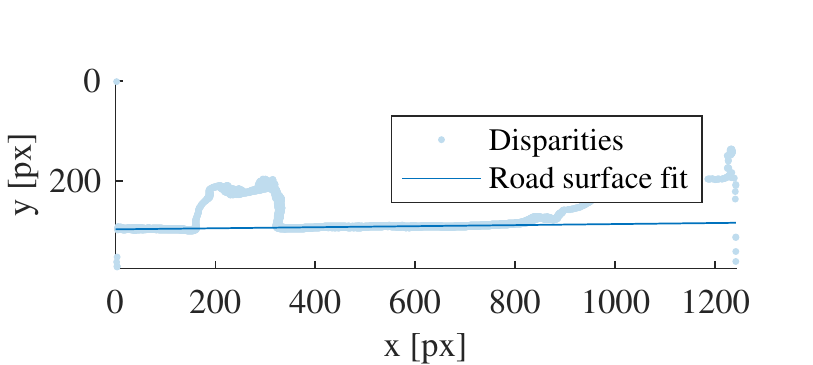}
    \caption{The camera roll angle is measured by sampling all pixels with disparities between 0.03 and 0.031 from the disparity map. A Hough line detector is then used to fit the road surface.}
    \label{fig:roll_fit}
\end{figure}

\begin{figure}
    \centering
    \includegraphics{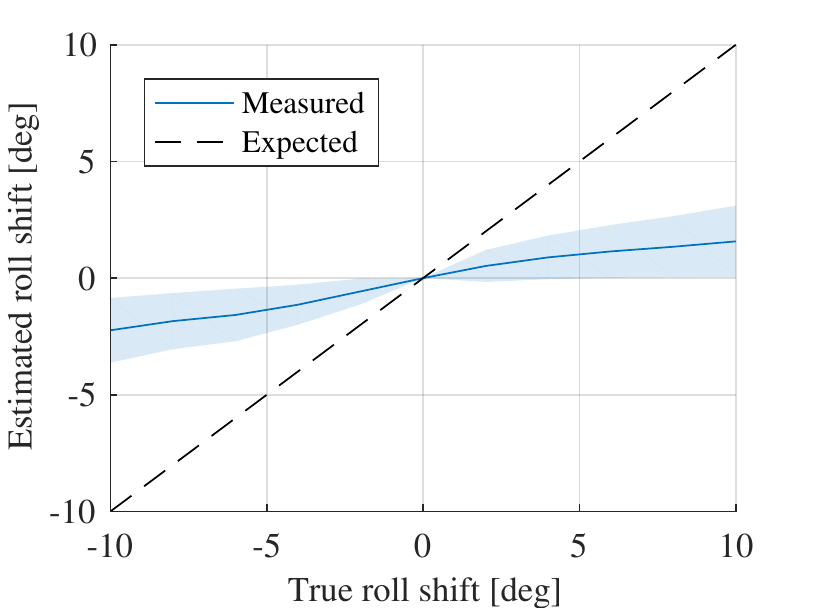}
    \caption{True and estimated roll shifts in the cropped images. The change in road surface angle is smaller than the true angle at which the images are cropped. Shaded regions indicate $\pm 1\,\text{SD}$ ($N=189$, eleven outliers $>3\,\text{SD}$ removed).}
    \label{fig:roll}
\end{figure}

Similarly to the pitch angle, the roll angle of the camera influences the depth estimate towards obstacles. If the camera has a nonzero roll angle, the distance towards obstacles does not only depend on their vertical position but also on their horizontal position in the image.

A similar experiment was performed as for the pitch angle: a smaller region of the images was cropped at varying angles (Figure \ref{fig:roll_crop}). The roll angle was then extracted from the estimated depth map (Figure \ref{fig:roll_fit}).
As in the previous experiment, we look for a correlation between the camera angle and the \emph{change} in the estimated angle of the road surface.
The results is shown in Figure \ref{fig:roll} and is similar to that for pitch angles: \monodepth\ is able to detect a roll angle for the camera, but this angle is too small in the resulting depth maps.



\section{Obstacle recognition}
\label{sec:appearance}

\begin{figure}
    \centering
    \includegraphics{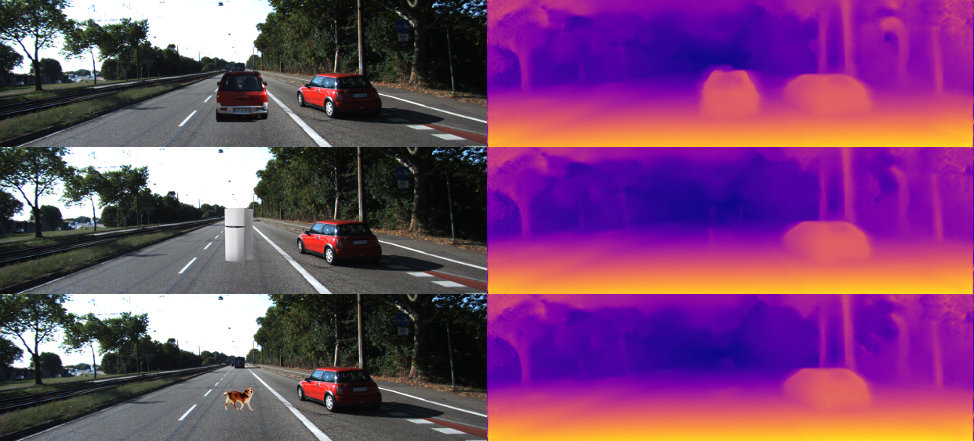}
    \caption{Objects that are not found in the training set (fridge, dog) are not reliably detected when pasted into the image.}
    \label{fig:other_objects}
\end{figure}

\begin{figure*} 
    \centering
    \includegraphics{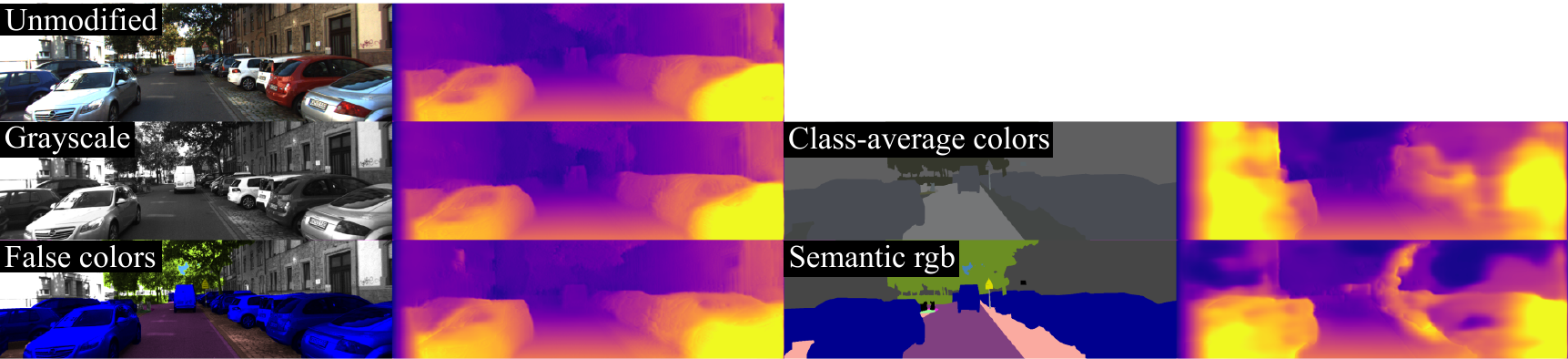}
    \caption{Example images and depth maps for unmodified, grayscale, false color, class average color, and semantic rgb images.}
    \label{fig:colorized}
\end{figure*}

\begin{table*}
    \centering
    \begin{tabular}{lcccccccc}
        \toprule
        Test set & Abs Rel & Sq Rel & RMSE & RMSE log & D1-all & $\delta<1.25$ & $\delta<1.25^2$ & $\delta<1.25^3$ \\
        \midrule
        Unmodified images & \textbf{0.124} & 1.388 & \textbf{6.125} & \textbf{0.217} & \textbf{30.272} & \textbf{0.841} & \textbf{0.936} & \textbf{0.975} \\
        Grayscale & 0.130 & 1.457 & 6.350 & 0.227 & 31.975 & 0.831 & 0.930 & 0.972 \\
        False colors & 0.128 & \textbf{1.257} & 6.355 & 0.237 & 34.865 & 0.816 & 0.920 & 0.966 \\
        Semantic rgb & 0.192 & 2.784 & 8.531 & 0.349 & 46.317 & 0.714 & 0.850 & 0.918 \\
        Class-average colors & 0.244 & 4.159 & 9.392 & 0.367 & 50.003 & 0.691 & 0.835 & 0.910 \\
        \bottomrule
    \end{tabular}
    \caption{\monodepth's performance on images with disturbed colors or texture. 
    The unmodified image results were copied from \cite{Godard2017}; the table lists results without post-processing.
    Error values for images that keep the value channel intact (\emph{grayscale} and \emph{false colors}) are close to the unmodified values. Images where the value information is removed and the objects are replaced with flat colors (\emph{segmentation map}, \emph{class-average colors}) perform significantly worse.
    }
    \label{tab:colorized}
\end{table*}

\begin{figure}
    \centering
    \includegraphics{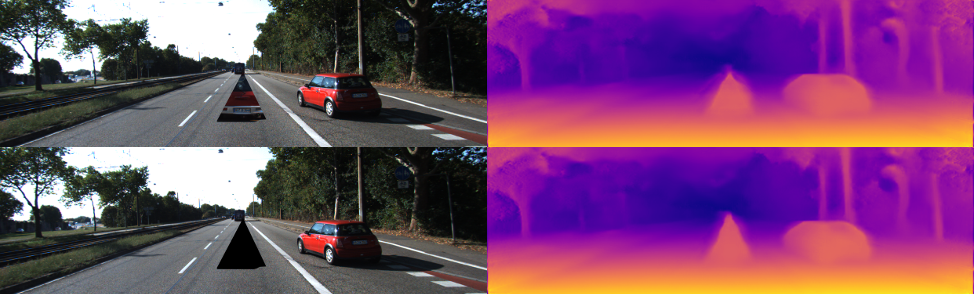}
    \caption{Objects do not need to have a familiar texture nor shape to be detected. The distance towards these non-existent obstacles appears to be determined by the position of their lower extent.}
    \label{fig:shape_texture}
\end{figure}

\begin{figure*} 
    \centering
    \includegraphics{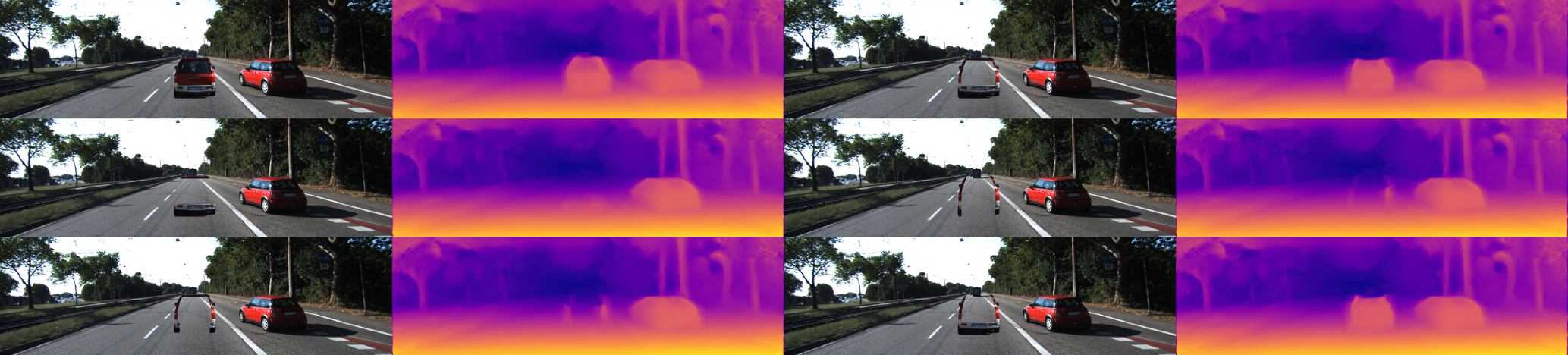}
    \caption{Influence of car parts and edges on the depth map. Removing the center of the car (top right) has no significant influence on the detection. The car's bottom and side edges (bottom right) seem most influential for the detected shape, which is almost identical to the full car image (top left).}
    \label{fig:edges}
\end{figure*}

Section \ref{sec:pos_vs_scale} has shown that \monodepth\ uses the vertical position of objects in the image to estimate their distance. 
The only knowledge that is required for this estimate is the location of the object's ground contact point. Since no other knowledge about the obstacle is required (e.g. its real-world size), this suggests that \monodepth\ can estimate the distance towards arbitrary obstacles. However, Figure \ref{fig:other_objects} shows that this is not always the case. The car is recognized as an obstacle, but the other objects are not recognized and appear in the depth map as a flat road surface.

In order to correctly estimate the depth of an obstacle, the neural network should be able to: 1) find the ground contact point of the obstacle, as this is used to estimate its distance, and 2) find the outline of the obstacle, in order to fill in the corresponding region in the depth map. The results of Figure \ref{fig:other_objects} suggest that \monodepth\ relies on a set of features that are applicable to cars but not to the other objects inserted into the test images. 

\subsection{Color and Texture}\label{sec:color}

The objects inserted in Figure \ref{fig:other_objects} differ from cars in terms of color, texture and shape.
In a first experiment, we investigate how color and texture affect \monodepth's performance by evaluating it on modified versions of the KITTI images. To investigate the influence of color, two new test sets are created: one in which the images are converted to grayscale to remove all color information, and one in which the hue and saturation channels are replaced by those from KITTI's semantic\_rgb dataset to further disturb the color information in the image. Two other datasets are used to test the influence of texture: a dataset in which all objects are replaced by a flat color that is the average of that object class -- removing all texture but keeping the color intact -- and the semantic\_rgb set itself where objects are replaced with unrealistic flat colors. Examples of the modified images and resulting depth maps are shown in Figure \ref{fig:colorized}, the performance measures are listed in Table \ref{tab:colorized}.

As long as the value information in the images remains unchanged (the \emph{unmodified}, \emph{grayscale} and \emph{false color} images), \monodepth's performance  remains roughly the same. A slight increase in error is observed for grayscale images and a further increase when false colors are added, but all error and performance measures remain close to those from the original test set.
This suggests that the exact color of obstacles does not strongly affect the depth estimate.

When only flat colors are used (\emph{class-average colors} and \emph{semantic rgb}) a noticeable drop in performance is observed.
The network also performs better on the semantic rgb dataset with false colors than on the class-average colors dataset where the colors are more realistic, which further suggests that the exact color of obstacles does not matter. Instead, other features such as the contrast between adjacent regions or bright and dark regions within objects may be more important.

\subsection{Shape and contrast}
Objects do not need a familiar texture nor shape to be detected (Figure \ref{fig:shape_texture}). The triangles pasted into the image are properly detected, where the top one has a car texture and the bottom is uniformly black.
The distances seem to follow from the lowest point of these unfamiliar objects. This supports the claim that distance is estimated based on vertical position in the image with the assumption that objects rest on the ground.
 
In particular, the edges of an object and their contrast with the environment seem to matter most for correct distance estimation. In Figure \ref{fig:edges}, we have removed various parts of the car. The resulting depth maps suggest that the neural network `fills in' an object, mainly based on its bottom and side edges. For instance, the bottom right image, with only the bottom and side edges, still leads to an almost fully detected car. Leaving out the bottom edge leads to the detection of the car sides as two separate, thin objects. 

\begin{figure}
    \centering
    \includegraphics{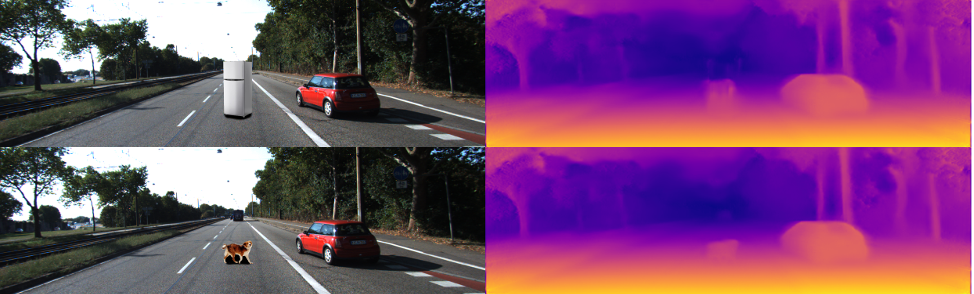}
    \caption{Adding a shadow at the bottom of the objects of Figure \ref{fig:other_objects} causes them to be detected. The fridge, however, is only detected up to the next horizontal edge between the bottom and top doors.}
    \label{fig:shadow}
\end{figure}

Note that the edges in Figure \ref{fig:edges} have significant contrast with the environment.
In the KITTI dataset, the shadows underneath the cars provide a highly contrasting bottom edge in nearly all cases.
Indeed, adding a similar shadow at the bottom of the undetected objects from Figure \ref{fig:other_objects} leads to their successful detection (Figure \ref{fig:shadow}). Still, in the depth map the fridge only extends to the edge between the fridge's two compartments, making it vertically smaller than it should be. 




\section{Conclusions and future work}\label{sec:discussion}


In this work we have shown that the \monodepth\ neural network by Godard \etal \cite{Godard2017} primarily uses the vertical position of objects in the image to estimate their depth, rather than their apparent size.
This estimate depends on the pose of the camera, but changes to this pose are not completely accounted for, leading to an under- or overestimation of the distance towards obstacles when the camera pitch changes.
We further show that \monodepth\ can detect objects that do not appear in the training set, but that this detection is not always reliable and depends on external circumstances such as the presence of a shadow under the object.
These are important limitations that were not mentioned in the original work -- nor in any other work on monocular depth estimation -- thereby highlighting our point that the learned behavior of these networks requires more attention than it is currently receiving.


The results in this work are based on a single neural network (\monodepth) trained on only one dataset (KITTI). We do not know yet whether other neural networks for monocular depth estimation behave in the same manner. In future work we plan to repeat the experiments published here with other networks and datasets to see whether the same behavior is learned.

{\small
\bibliographystyle{ieee}
\bibliography{library,misc}

\begin{thebibliography}{10}\itemsep=-1pt

\bibitem{Brenner2018}
E.~Brenner and J.~B. Smeets.
\newblock {Depth Perception}.
\newblock In J.~Wixted, editor, {\em Stevens' Handbook of Experimental
  Psychology and Cognitive Neuroscience}, chapter Depth Perc, pages 385--414.
  John Wiley {\&} Sons, New York, 4 edition, 2018.

\bibitem{Cutting1995}
J.~E. Cutting and P.~M. Vishton.
\newblock {Perceiving Layout and Knowing Distances: The Integration, Relative
  Potency, and Contextual Use of Different Information about Depth}.
\newblock In {\em Perception of Space and Motion}, pages 69--117. Elsevier,
  1995.

\bibitem{Eigen2014}
D.~Eigen, C.~Puhrsch, and R.~Fergus.
\newblock {Depth Map Prediction from a Single Image using a Multi-Scale Deep
  Network}.
\newblock In {\em Advances in Neural Information Processing Systems 27}, pages
  2366--2374. Curran Associates, Inc., 2014.

\bibitem{Epstein1966a}
W.~Epstein.
\newblock {Perceived Depth as a Function of Relative Height under Three
  Background Conditions}.
\newblock {\em Journal of Experimental Psychology}, 72(3):335--338, 1966.

\bibitem{Erhan2009}
D.~Erhan, Y.~Bengio, A.~Courville, and P.~Vincent.
\newblock {Visualizing higher-layer features of a deep network}, 2009.

\bibitem{Gardner2010}
J.~S. Gardner, J.~L. Austerweil, and S.~E. Palmer.
\newblock {Vertical position as a cue to pictorial depth: Height in the picture
  plane versus distance to the horizon}.
\newblock {\em Attention, Perception, {\&} Psychophysics}, 72(2):445--453,
  2010.

\bibitem{Garg2016}
R.~Garg, V.~B. Kumar, G.~Carneiro, and I.~Reid.
\newblock {Unsupervised CNN for Single View Depth Estimation: Geometry to the
  Rescue}.
\newblock In B.~Leibe, J.~Matas, N.~Sebe, and M.~Welling, editors, {\em
  European Conference on Computer Vision}, pages 740--756, Cham, 2016. Springer
  International Publishing.

\bibitem{Gibson1950}
J.~J. Gibson.
\newblock {\em {The perception of the visual world}}.
\newblock Houghton Mifflin, Oxford, England, 1950.

\bibitem{Godard2017_github}
C.~Godard.
\newblock {monodepth: Unsupervised single image depth prediction with CNNs}.
\newblock {\em https://github.com/mrharicot/monodepth}, Apr 2017.

\bibitem{Godard2017}
C.~Godard, O.~{Mac Aodha}, and G.~J. Brostow.
\newblock {Unsupervised Monocular Depth Estimation with Left-Right
  Consistency}.
\newblock {\em The IEEE Conference on Computer Vision and Pattern Recognition
  (CVPR)}, 2017.

\bibitem{Menze2015}
M.~Menze and A.~Geiger.
\newblock {Object scene flow for autonomous vehicles}.
\newblock {\em Proceedings of the IEEE Computer Society Conference on Computer
  Vision and Pattern Recognition}, 07-12-June:3061--3070, 2015.

\bibitem{Olah2017a}
C.~Olah, A.~Mordvintsev, and L.~Schubert.
\newblock {Feature Visualization}.
\newblock {\em Distill}, 2017.

\bibitem{Ooi2001}
T.~L. Ooi, B.~Wu, and Z.~J. He.
\newblock {Distance determined by the angular declination below the horizon}.
\newblock {\em Nature}, 414:197--200, 2001.

\bibitem{Saxena2006}
A.~Saxena, S.~H. Chung, and A.~Y. Ng.
\newblock {Learning Depth from Single Monocular Images}.
\newblock {\em Advances in Neural Information Processing Systems},
  18:1161--1168, 2006.

\bibitem{Silberman2012}
N.~Silberman, D.~Hoiem, P.~Kohli, and R.~Fergus.
\newblock {Indoor Segmentation and Support Inference from RGBD Images}.
\newblock In A.~Fitzgibbon, S.~Lazebnik, P.~Perona, Y.~Sato, and C.~Schmid,
  editors, {\em Computer Vision -- ECCV 2012}, pages 746--760, Berlin,
  Heidelberg, 2012. Springer Berlin Heidelberg.

\bibitem{Sousa2012}
R.~Sousa, J.~B. Smeets, and E.~Brenner.
\newblock {Does size matter?}
\newblock {\em Perception}, 41(12):1532--1534, 2012.

\bibitem{Szegedy2013}
C.~Szegedy, W.~Zaremba, I.~Sutskever, J.~Bruna, D.~Erhan, I.~Goodfellow, and
  R.~Fergus.
\newblock {Intriguing properties of neural networks}.
\newblock {\em arXiv preprint arXiv:1312.6199}, 2013.

\bibitem{Zeiler2013}
M.~D. Zeiler and R.~Fergus.
\newblock {Visualizing and Understanding Convolutional Networks}.
\newblock In D.~Fleet, T.~Pajdla, B.~Schiele, and T.~Tuytelaars, editors, {\em
  Computer Vision -- ECCV 2014}, pages 818--833, Cham, 2014. Springer
  International Publishing.

\end{thebibliography}
}

\end{document}